\documentclass[10pt,twocolumn,letterpaper]{article}
\pdfoutput=1
\usepackage{cvpr}
\usepackage{times}
\usepackage{epsfig}
\usepackage{graphicx}
\usepackage{amsmath}
\usepackage{amssymb}
\usepackage{multirow}
\usepackage[dvipsnames]{xcolor}
\usepackage{booktabs}
\usepackage{float}
\usepackage{algorithm}
\usepackage[noend]{algpseudocode}

\usepackage[pagebackref=true,breaklinks=true,letterpaper=true,colorlinks,bookmarks=false]{hyperref}

\cvprfinalcopy % *** Uncomment this line for the final submission

\def\cvprPaperID{166} % *** Enter the CVPR Paper ID here

\ifcvprfinal\fi

\begin{document}

%%%%%%%%% TITLE
\title{Guess Where? Actor-Supervision for Spatiotemporal Action Localization}

\author{Victor Escorcia$^{1*}$ \and Cuong D. Dao$^{1}$ \and Mihir Jain$^3$ \and Bernard Ghanem$^1$ \and Cees Snoek$^{2*}$ \and
KAUST$^1$, University of Amsterdam$^2$, Qualcomm Technologies, Inc.$^3$ \\
% \and
}

\maketitle
%\thispagestyle{empty}

%%%%%%%%% ABSTRACT
\begin{abstract}
This paper addresses the problem of spatiotemporal localization of actions in videos. Compared to leading approaches, which all learn to localize based on carefully annotated boxes on training video frames, we adhere to a weakly-supervised solution that only requires a video class label. We introduce an actor-supervised architecture that exploits the inherent compositionality of actions in terms of actor transformations, to localize actions. We make two contributions. First, we propose actor proposals derived from a detector for human and non-human actors intended for images, which is linked over time by Siamese similarity matching to account for actor deformations. Second, we propose an actor-based attention mechanism that enables the localization of the actions from action class labels and actor proposals and is end-to-end trainable. Experiments on three human and non-human action datasets show actor supervision is state-of-the-art for weakly-supervised action localization and is even competitive to some fully-supervised alternatives.
\end{abstract}
\let\thefootnote\relax\footnotetext{$^*$ Work done during VE's internship at Qualcomm Technologies, Inc.}

%------------------------------------------------------------------------
%%%%%%%%%%%%%%%%%%%%%%%%%%%%%%%%%%%%%%%%%%%%%%%%%%%%%%%%%%%%%%%%%%%%%%%%%
%%%%%%%%%%%%%%%%%%%%%%%%%%%%%%%%%%%%%%%%%%%%%%%%%%%%%%%%%%%%%%%%%%%%%%%%%
%------------------------------------------------------------------------
%--------------------------------------
\section{Introduction}
%--------------------------------------
The goal of this paper is to localize and classify actions like \textit{skateboarding} or \textit{walking with dog} in video by means of its enclosing spatiotemporal tube. Empowered by action proposals \cite{Jain_cvpr14, WeinzaepfelHS15,ZhuVL17}, deep learning \cite{gkioxariCVPR15actionTubes, Saha_bmvc16, kalogeiton17iccv, HouCS_iccv17} and carefully labeled datasets containing spatiotemporal annotations \cite{ucf101,RodriguezCVPR2008,XuHsXiCVPR2015}, progress on this challenging topic has been considerable. However, the dependence on deep learning and spatiotemporal boxes is also hampering further progress, as annotating tubes inside video is tedious, costly and error prone~\cite{MettesECCV16}. We strive for action localization founded on deep learning without the need for spatiotemporal video supervision during training.

Others have also considered weakly-supervised action localization \cite{SivaBMVC11,MettesSC17,LiGJS16}. Siva and Xiang \cite{SivaBMVC11} detect possible human actor locations and sample a set of action cuboids in its spatiotemporal neighborhood. A multiple instance learning algorithm then exploits inter- and intra-class distances between action class labels to find the optimal cuboid. Mettes \etal \cite{MettesSC17} also rely on multiple instance learning, but rather than cuboids, they start from hundreds of unsupervised action proposals \cite{Gemert_BMVC15}. To guide the selection of the best proposal, this work combines information from multiple cues such as detected human actors with action-specific video labels. We also rely on human actor detectors to alleviate the need for spatiotemporal annotations. Different from \cite{MettesSC17}, we do not use them to select from among a set of proposals \textit{a posteriori}, but rather exploit them to define the proposals \textit{a priori}. Different from \cite{SivaBMVC11}, we do not rely on cuboid representations and multiple instance learning, but rather adopt an end-to-end trainable neural network approach. Li \etal \cite{LiGJS16} also propose an end-to-end neural network tailored towards action localization. Their VideoLSTM learns to attend to spatiotemporal regions of interest based on action class labels. We also rely on an attention mechanism, but rather than selecting individual pixels, we prefer to select the most suited proposals. Moreover, our attention mechanism is not only derived from the action class, but it also considers the human (or non-human) actor.

\begin{figure}[!t]
\label{fig:pull-figure}
\centering
  \includegraphics[width=0.48\textwidth]{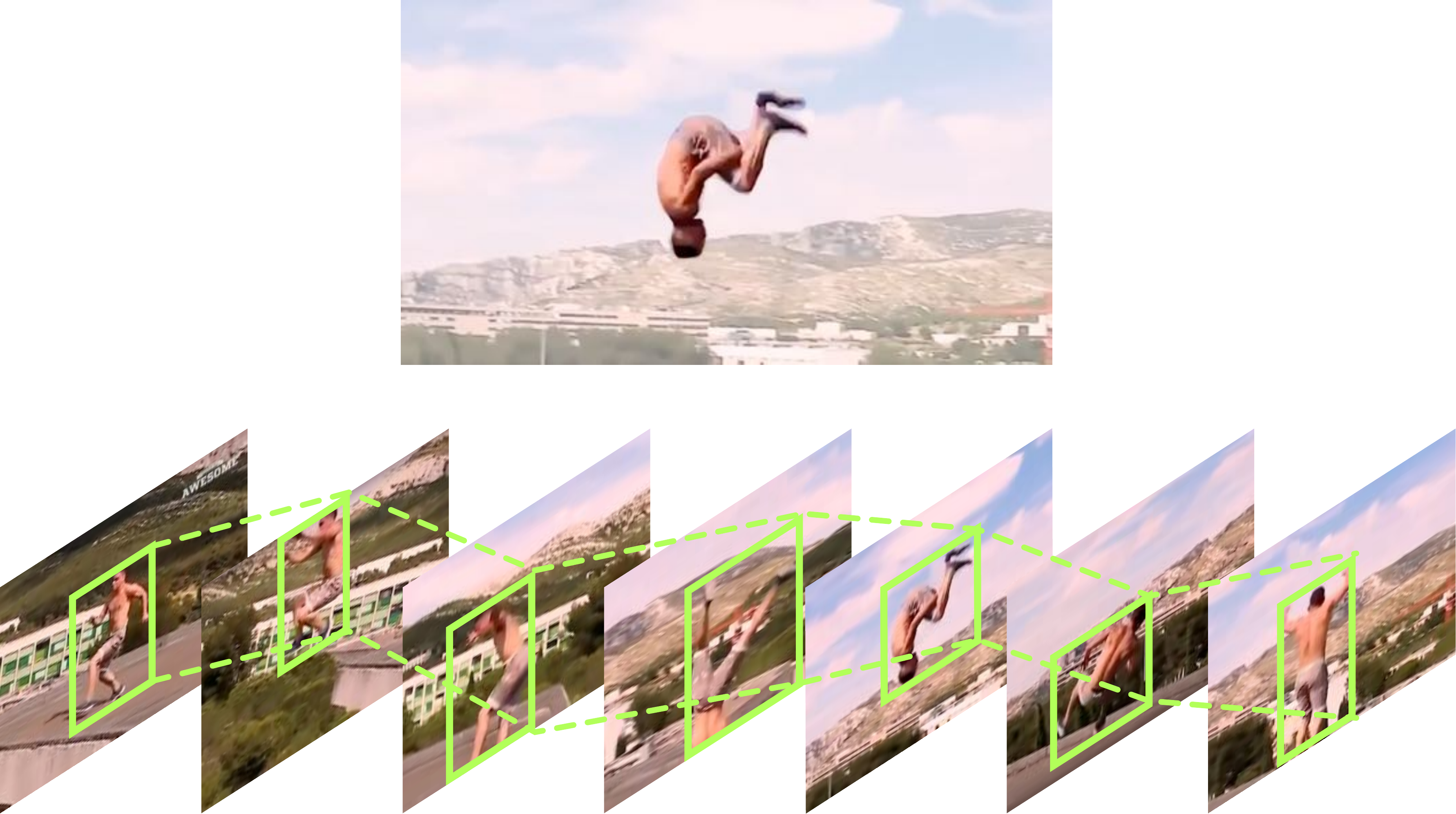}
\caption{We propose actor-supervision as a means for weakly-supervised spatiotemporal action localization in the video. Our method relies only on action labels at the video level during the learning stage.}
\vspace{-3mm}
\end{figure}

We propose the notion of actor-supervision as a means for weakly-supervised spatiotemporal action localization. It exploits the inherent compositionality of actions in terms of actor transformations, to localize actions without using spatiotemporal annotations of the training videos. 
We make two technical contributions in this work. First, we introduce actor proposals; a means to generate candidate tubes that are likely to contain an action and that do not require any action video annotations for training. Instead, we derive our proposals from a detector for human and non-human actors, intended for images, combined with Siamese similarity matching to account for actor deformations over time. Second, we introduce an actor-based attention mechanism that allows the localization of actions only from action labels and can be trained in an end-to-end fashion with stochastic gradient descent. Experiments on three human and non-human action datasets show that our actor supervision is the state-of-the-art for weakly-supervised spatiotemporal action localization and is even competitive with some fully-supervised alternatives.

%------------------------------------------------------------------------
%%%%%%%%%%%%%%%%%%%%%%%%%%%%%%%%%%%%%%%%%%%%%%%%%%%%%%%%%%%%%%%%%%%%%%%%%
%%%%%%%%%%%%%%%%%%%%%%%%%%%%%%%%%%%%%%%%%%%%%%%%%%%%%%%%%%%%%%%%%%%%%%%%%
%------------------------------------------------------------------------
%=======================================
\section{Related work}
%=======================================
Typical approaches for action localization first generate spatiotemporal action proposals and then classify them with the appropriate action label. We discuss work related to these two aspects of action localization and group them by the amount of supervision needed.
%---------------------------------------
\subsection{Action proposals}
%---------------------------------------
\paragraph{Supervised action proposals} generate box proposals and classify them per action for each individual frame. In addition to video-level class labels, bounding-box ground-truth for each action instance across all the frames is required. In~\cite{gkioxariCVPR15actionTubes,WeinzaepfelHS15}, the box proposals come from object proposals~\cite{Jasper:selective,Zitnick2014}, and a two-stream CNN is learned to classify these boxes into action classes. More recently,~\cite{kalogeiton17iccv,Saha_iccv17_AMTnet,Saha_bmvc16,GSingh_iccv17_online} action boxes are generated by two-stream extensions of modern object detectors~\cite{SSD_eccv16,faster_rcnn_nips15},
like SSD~\cite{GSingh_iccv17_online} and its extension~\cite{kalogeiton17iccv}. Zhu \etal \cite{ZhuVL17} introduced a spatiotemporal convolutional regression network for action box generation. For all these works, once the action boxes per frames are established they are linked together to create action proposals per video. The common tactic is to exploit dynamic programming based on the Viterbi alogrthm~\cite{gkioxariCVPR15actionTubes}.
\vspace{-4mm}
\paragraph{Unsupervised action proposals} do not require any class labels or bounding box ground-truth. The traditional sliding-window sampling is unsupervised but has an exponentially large search space. More efficient methods sample boxes from super-voxels to generate action proposals~\cite{Jain_cvpr14,Jain2017,oneataECCV14spatemprops}. Clustering of motion trajectories has also proven to be an effictive choice~\cite{Gemert_BMVC15,chen2015action}. Puscas \etal~\cite{marianICCV2015unsupervisedTube} use box proposals~\cite{Jasper:selective} but, unlike supervised methods, link them by optical flow and dense trajectory based matching. 
\vspace{-4mm}
\paragraph{Weakly-supervised action proposals} do not use per frame box-level ground-truth for all the video frames~\cite{klaser2012human,tian_iccv11,Tran:nips12,yuCVPR15fasttubes}. Instead, they rely on object detectors trained on images, to get detections. Among these \cite{klaser2012human} and \cite{yuCVPR15fasttubes} are the closest to our approach. Yu and Yuan~\cite{yuCVPR15fasttubes} use a human detector and motion scores to locate boxes and compute an actionness score for each of them. The boxes are then linked by formulating it as a maximum set convergence problem. We instead use a similarity based tracker, as our objective is to handle deformation, motion cannot do that. Kl\"aser \etal~\cite{klaser2012human} uses an upper-body detector per frame and tracks them by optical flow feature points to generate spatiotemporal tubes. Again, tracking matches optical flow feature points, while we prefer Siamese similarity matching~\cite{tao2016sint,bertinetto2016fully} that is more robust to deformations and can recover from loose and imprecise detection boxes.

Full supervision results in more precise boxes but scales poorly as the number of action classes grows. Unsupervised proposals are more scalable but boxes are often less precise. Our approach achieves the best of both worlds. We obtain box precision by using a human and non-human actor detector and then link the boxes from consecutive frames by Siamese matching, making them robust to deformations. At the same time, our approach is action-class agnostic and hence more scalable.
%---------------------------------------
\subsection{Proposal classification}
%---------------------------------------
\paragraph{Supervised classification} is the default approach in the action localization literature. Methods train classifiers using box-level ground-truth for each action class and apply it on each action proposal for each test video, \eg~\cite{gkioxariCVPR15actionTubes,kalogeiton17iccv,Saha_iccv17_AMTnet,Saha_bmvc16,GSingh_iccv17_online, WeinzaepfelHS15}. Others, who relied on unsupervised or weakly-supervised action proposals, now train their action proposal classifiers in a supervised fashion using bounding-box ground-truth across frames~\cite{Jain2017,Gemert_BMVC15,chen2015action}. 
\vspace{-4mm}
\paragraph{Unsupervised and zero-shot classification} has been addressed as well.
Puscas \etal~\cite{marianICCV2015unsupervisedTube} classify their unsupervised proposals using tube-specific, class agnostic detectors, trained in a 2-stage transductive learning framework, to extract the final tubes. Another fully unsupervised action localization approach is proposed by Soomro and Shah~\cite{SoomroS17}. It starts with supervoxel segmentation, automatically discovers action classes by using discriminative clustering and localizes actions by knapsack optimization for supervoxel selection. In \cite{Jain15iccv}, action proposals are classified in a zero-shot fashion by encoding them into a semantic word embedding spanned by object classifiers. A spatial-aware object embedding is proposed in \cite{MettesICCV17}, that also captures actors, relevant objects and their spatial realtions. All the training happens on images and text, no videos are needed.

\begin{figure*}[ht]
\begin{center}
\includegraphics[width=0.95\textwidth]{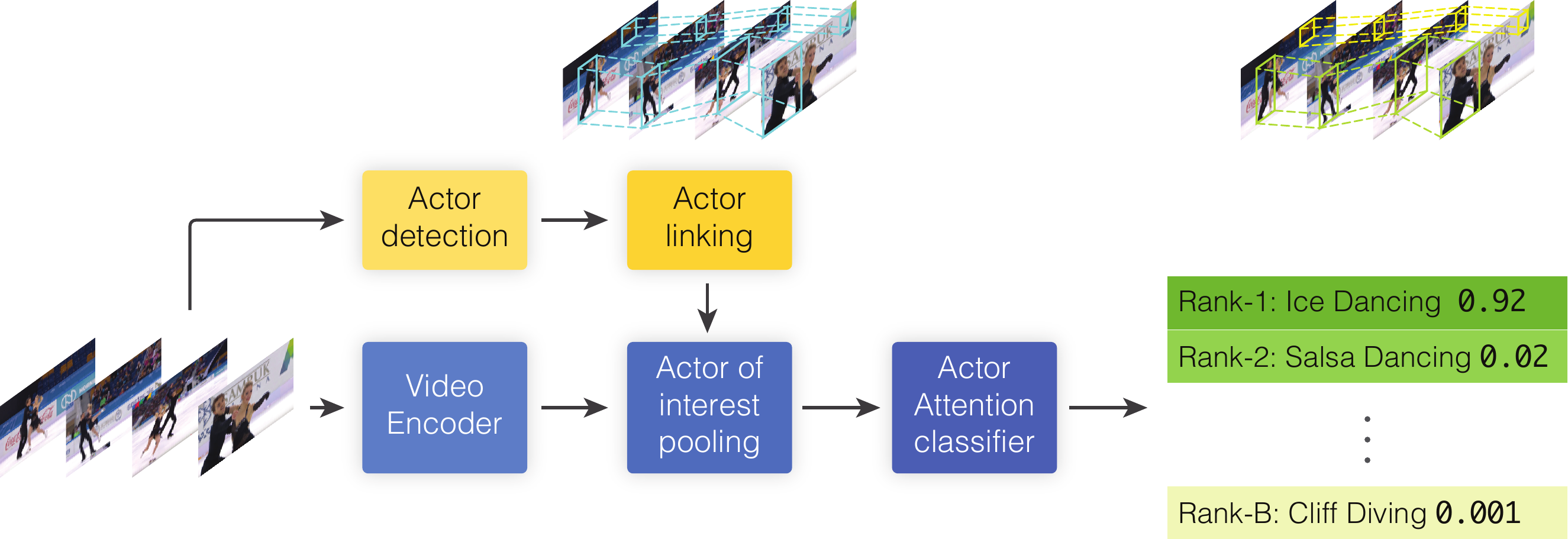}
\end{center}
\vspace{-10pt}
\caption{We employ actor-supervision to design a deep architecture for spatiotemporal action localization. To this end, we infuse the notion of actors in our architecture. The top stream corresponds to our actor proposal pillar which generates relevant actor tubes where is likely to find actions in the videos. The bottom stream illustrates our actor attention which classifies the action occuring on each actor tube and sorts them based on the relevance for the predicted action class.}
\label{fig:pipeline}
\vspace{-10pt}
\end{figure*}

\vspace{-4mm}
\paragraph{Weakly-supervised classification} refrains from using per frame box-level ground-truth for classifying action proposals. A considerable reduction in annotation effort may be achieved by replacing box annotations with point annotations and unsupervised action proposals \cite{MettesECCV16}, but it still demands manual labor. An alternative is to rely on human body parts \cite{MaZIS13} or human detectors trained on image benchmarks \cite{RussakovskyDSKS15,LinMBHPRDZ14} to steer the localization in video; either by defining the search space of most likely action locations \cite{SivaBMVC11} or by selecting the most promising (unsupervised) action proposal \cite{MettesSC17}. 
Apart from human actor detectors, attention mechanisms in deep neural network architectures have been explored. Sharma \etal \cite{sharma2015action} and Li \etal \cite{LiGJS16} show that such a mechanism provides the action location for free when training on action class labels only.
We also rely on human (and non-human) actor detectors but exploit them to generate a limited set of actor proposals, from which we select the best ones per action, based on an actor attention mechanism that only requires action class labels. Without the need for box annotations per video frame, we achieve results not far behind the supervised methods and much better than unsupervised methods.

%------------------------------------------------------------------------
%%%%%%%%%%%%%%%%%%%%%%%%%%%%%%%%%%%%%%%%%%%%%%%%%%%%%%%%%%%%%%%%%%%%%%%%%
%%%%%%%%%%%%%%%%%%%%%%%%%%%%%%%%%%%%%%%%%%%%%%%%%%%%%%%%%%%%%%%%%%%%%%%%%
%------------------------------------------------------------------------
\section{Proposed Actor-Supervised Architecture}
Given a video clip, we aim for the spatiotemporal localizaton of an action. Contrary to the expensive and error-prone approaches of using action supervision, typically in the form of a set of action boxes for individual frames throughout the video, our approach learns to perform this task based on action class labels at the video level only.

To deal with the inherent difficulty of the problem, we introduce actor supervision to take advantage of the fact that actors are precursors of actions.  Actions result from an actor going through certain transformation, while possibly interacting with other actors and/or objects in the process.  This means that actors not only locate the action in the video, but also one can learn to rank the potential actor locations for a given action class. Based on these premises, we design a novel deep architecture for spatiotemporal action localization guided by actor supervision.

Figure \ref{fig:pipeline} illustrates the two pillars of our architecture, namely \textit{actor proposals} and \textit{actor attention}.
In a nutshell, our approach enables the localization of actions with minimal supervision by (i) embedding the concept of actors in the architecture; (ii) exploiting existing knowledge and progress in object detection and object tracking; and (iii) introducing a powerful attention mechanism suitable for learning a meaningful representation of actions. In the following subsections we disclose full details of each pillar.
\subsection{Actor proposals}
\begin{figure*}[!th]
\centering
    \includegraphics[width=0.90\textwidth]{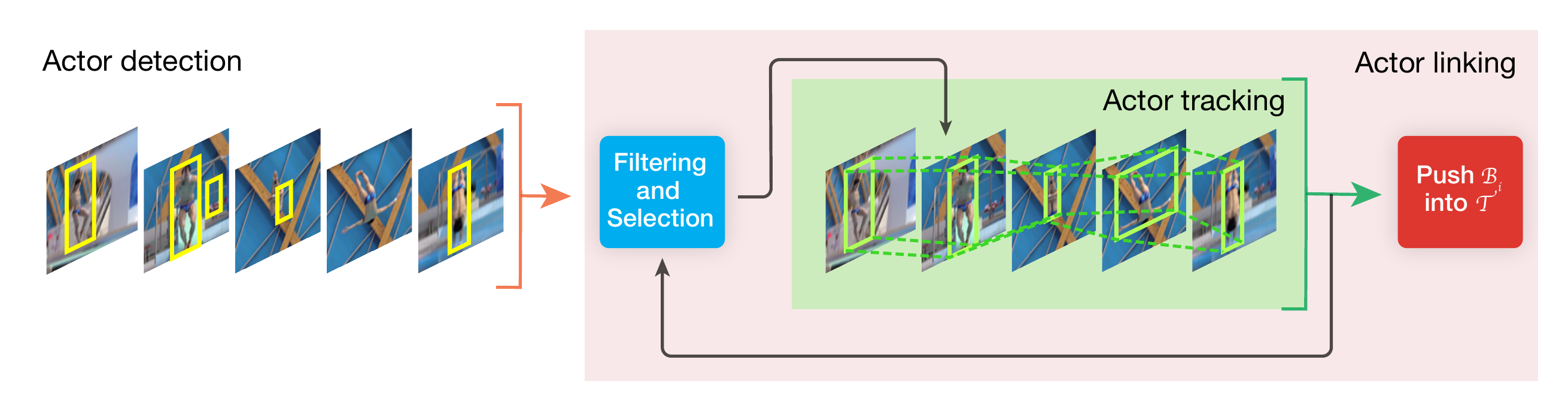}
\caption{We generate actor proposals by detecting the most likely actor locations with the aid of an object detector. Our actor linking module selects the most relevant detections and carefully tracks them throughout the video using a Siamese Network, which robustly overcomes actor deformations typical during the performance of the action. Additionally, we filter out detections with high similarity with the boxes of the tubes.
}
\label{fig:actor-proposals}
\vspace{-3mm}
\end{figure*}

%sequence of frames appropriate for a weakly supervised setup.
Our actor proposals receive a video stream and generate a set of tubes, parametrized as a sequence of boxes $\mathcal{T} = \{\mathcal{B}_i\}$. The tubes outline the most likely spatiotemporal regions where an action may occur based on the presence of an actor.
It contains two modules: actor detection and actor linking, as shown in Figure \ref{fig:actor-proposals} and detailed next.

\vspace{-5mm}
\paragraph{Actor detection.} This module generates spatial locations where the actor of interest appears in the video. Respecting the requirements of our setup, this module adopts a pre-trained conv-net for object detection which predicts bounding boxes over all the frames of the video.
Despite the huge progress in object detection, the predictions are still imperfect due to false positive errors or missed detections of the actor. Missed detections typically occur when the actor undergoes a significant deformation, which is common in actions. For example, when performing a cartwheel in floor gymnastics, the shape of the actor changes when he/she is flipping upside down. In these cases, actor deformations, characteristic of the performance of the action, may involve signicant visual changes that do not fit the canonical model of the object category of the actor.

\vspace{-12pt}
\paragraph{Actor linking.} This stage carefully propagates the predictions of our actor detector throughout the video to generate an actor proposal tube. It complements the detector by filling the gaps left during the performance of the action, without demanding any annotation to tune the detector. In this way, our module is more robust to missed detections and consistenly retrieves complete actor tubes associated with actors.
We attain this goal with the aid of a robust similarity-based tracker along with a scheme to filter and select the boxes enabling detection boxes and tracker coordination. 

The similarity tracker exploits the temporal coherence between neighboring frames in the video, generating a box-sequence for every given box.
In practice, we employ a pre-trained similarity function learned by a Siamese network which strengthens the matching of the actor between a small neighborhood in adjacent frames. Once it is learned, this similarity function is transferable and remains robust against deformations of the actor \cite{tao2016sint,bertinetto2016fully}.

The filtering and selection scheme selects the best scoring detection boxes and sequentially feeds them to the tracker, which propogates them into box-sequences $\mathcal{B}_i$, also called tubes. This scheme also filters out the candidate detections similar to the boxes generated by the tracker 
reducing the amount of computation required. 
Section \ref{exp:boring-details} describes the implementation details about the conv-net architectures used to generate our actor proposals.
\subsection{Actor attention}
The second pillar of our approach is responsible for assigning action labels to the actor proposals in a weakly-supervised setting. It takes into account the visual appereance inside the actor proposals, and scores them based on the model for action classification, \ie models trained on video-level class labels only. The outcome of this module is a set of ranked proposals where it is likely to find particular actions in the video.

Figure \ref{fig:pipeline} illustrates the inner components of our actor attention which are detailed next. Briefly, this module combines recent advances in modern deep learning based video representations with an appropriate attention mechanism which enables it to learn without the aid of bounding boxes.

\vspace{3pt}
\noindent\textbf{Video encoder.~} The encoder transforms the video stream into a suitable space where our attention module can discern among different actions. In practice, we use a conv-net, which encodes video frames as response maps that also comprise spatial information.
Without loss of generality, an input video with $N$ frames and shape $N \times 3 \times W \times H$ produces a tensor of shape $N \times C \times W' \times H'$, where $C$ is the number of response maps in the last layer of the video encoder. $W'$ and $H'$ correspond to scaled versions of the original width and height, respectively, due to the pooling layers or convolutions with long stride.
\vspace{-5mm}
\paragraph{Actor of interest pooling.} The pooling takes as input the response maps from the video encoder and the set of actor proposals, and outputs a fixed size representation for each actor proposal.
This module identifies the regions associated with each actor proposal in the response maps, and extracts a smooth representation. This operation resembles an extension of the ROI pooling operation over proposals in space and time \cite{girshickICCV15fastrcnn}. In practice, we implemented this operation as the temporal average pooling over a bilinear interpolation between the box sequence conforming an actor proposal and the input feature map \cite{JaderbergSZK15, densecap}.

Concretely, given an input feature map $U$ of shape $N\times C\times W'\times H'$ and a set of actor proposals of shape $P\times N \times 4$, which represents coordinates of the bounding boxes of $P$ actor proposals of length $N$. We interpolate the features of $U$ to produce an output feature map $V$ of shape $N\times P\times C\times X \times Y$ where $X, Y$ are the hyper-parameters representing the size of the desired output features for each actor box. For each actor box, we perform bilinear interpolation by projecting the bounding box onto the corresponding $U_{n,:,:,:}$ and computing a uniform sampling grid of size $X\times Y$, inside each actor box, associating each element of $V$ with real-valued coordinates into $U$. We obtain $V$ by convolving with a sampling kernel $k\left (d \right )=\max(0, 1 - | d |)$:
\begin{equation}
V_{n,p,c,i,j} = \sum_{i'=1}^{W}\sum_{j'=1}^{H}U_{n, c,i',j'}k\left ( i'- x_{n,p,i,j}\right ) k \left (j'-y_{n,p,i,j} \right )
\label{eq:wth}
\end{equation}
Finally, we average pool the contribution of all the output features belonging to the same actor proposal, which gives us a tensor of shape $P\times C\times X\times Y$ corresponding to the final output of our actor of interest pooling.
\vspace{-4mm}
\paragraph{Actor classification.} We classify each actor proposal according to a pre-defined set of action classes. This module learns to map the fixed size representation of each actor proposal into the space of actions of interest. In practice, we employ a fully connected layer where the number of outputs corresponds to the number of classes.

During training, the main challenge is to learn an appropriate mapping of the actor representation into the action space without the obligation of explicit annotations for each actor proposal. For this reason, we propose an attention mechanism over the actor proposals that boostraps the action labels at the video level. In this way, we encourage the network to learn the action classifier by focusing on the actors that contribute to an appropriate classification.

In this work, we explore the use of an attention mechanism based on top-$k$ selection. It encourages the selection of the $k$ most relevant actors per class that contribute to perform a correct classification. In practice, we choose the top-$k$ highest scores from the fully connected layer for each action category, and average them to form a single logit vector for each video. Subsequently, we apply a softmax activation on the logits of each video.
\vspace{-4mm}
\paragraph{Learning.} We train our actor attention using the cross-entropy loss between the output of the softmax and the video label.
It is relevant to highlight that we do not use any spatiotemporal information about the actions for learning the parameters of our model.

In practice, we fit the parameters of our actor attention employing backpropagation and stochastic gradient descent.
In the case of the top-$k$ selection module, we use a binary mask during the backpropagation representing the subgradients of the selection process.
A weight decay strength of $1 \times 10^{-4}$ is used in our actor classification module.

%------------------------------------------------------------------------
%%%%%%%%%%%%%%%%%%%%%%%%%%%%%%%%%%%%%%%%%%%%%%%%%%%%%%%%%%%%%%%%%%%%%%%%%
%%%%%%%%%%%%%%%%%%%%%%%%%%%%%%%%%%%%%%%%%%%%%%%%%%%%%%%%%%%%%%%%%%%%%%%%%
%------------------------------------------------------------------------
\section{Experiments}
\subsection{Datasets}
We validate our approach on three public benchmarks for spatiotemporal action localization in videos.
\vspace{-4mm}
\paragraph{UCF-Sports.} This dataset consists of 150 videos from TV sport channels representing 10 different action categories such as weightlifting, diving, golf-swing, \etc \cite{SoomroZ2014}. We employ the evaluation protocol established by \cite{tian_iccv11}, but without using the box annotations in the training set.
\vspace{-4mm}
\paragraph{THUMOS13.} This dataset incorportates untrimmed videos and multiple instances per video. It consists of a subset of 3,294 videos derived from UCF101 featuring 24 different action categories. We use the training and testing partition from split 1 of UCF101 for evaluating our approach \cite{ucf101}. Note that we do not rely on the spatiotemporal box annotations of the training set.
\vspace{-4mm}
\paragraph{Non-human actors dataset.} It is a public dataset comprising 3,782 videos collected to model the relationship between actors and actions in videos. This dataset considers actions, such as flying, jumping, climbing \etc, as performed by various actors, such as ball, cat, \etc We do not to use the spatiotemporal annotations of the training set.
\vspace{-4mm}
\paragraph{Evaluation.} Following the standard protocol for action localization, we use the intersection over union (IoU) to measure the degree to which a candidate tube is associated with a given spatiotemporal action ground-truth annotation. Depending on the task and dataset of interest, we report the result in terms of Recall or mean Average Precision (mAP).
\begin{figure*}[!th]
\centering
  \includegraphics[width=0.99\textwidth]{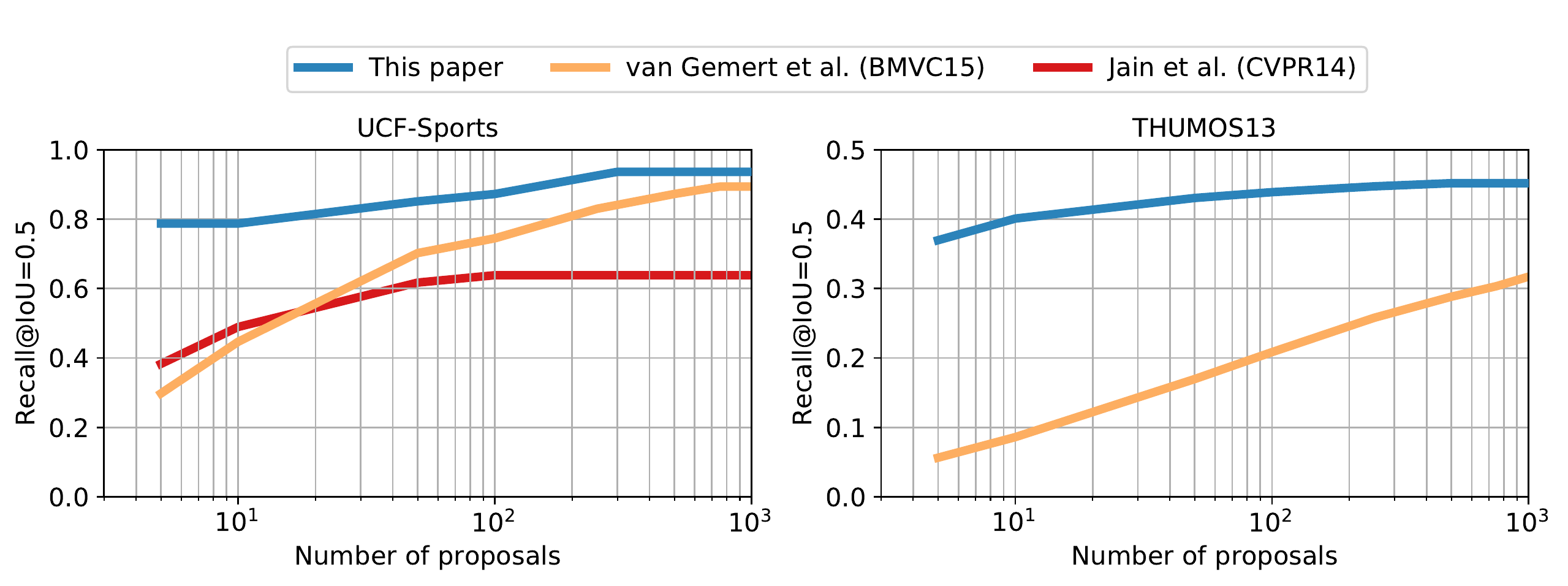}
\caption{Our actor proposals outperform previous unsupervised approaches for action proposals. We attribute its success to the use of actors as semantic information relevant for the grounding of the actions. Notably, our actor selection scheme retrieves enough relevant action tubes from a much smaller pool, which is advantageous in the context of retrieval and spatiotemporal localization of actions.}
\label{fig:proposal-comparison}
\vspace{-2mm}
\end{figure*}
%------------------------------------------------------------------------
%------------------------------------------------------------------------
\vspace{-4mm}
\subsection{Implementation details} \label{exp:boring-details}
\paragraph{Actor proposals.} We use a single-shot multi-box detector \cite{SSD_eccv16} to detect an actor of interest in every frame. We train the detector on all the categories of MS-COCO \cite{LinMBHPRDZ14} and limit the detections used to the actors of interest according to the action categories defined in each dataset. The base network of our actor detector is an InceptionV2 network pre-trained on ILSVCR-12 \cite{RussakovskyDSKS15}.
After computing all the detections, we only track the detections selected by our actor linking forward-and-backward over the entire video. For this purpose, we employ a multi-scale fully-convolutional Siamese-tracker \cite{bertinetto2016fully} trained on the ALOV dataset \cite{SmeuldersCCCDS14}. The base network of our tracker corresponds to the first four convolutional blocks of VGG-16. In our current setup, the actor selector ignores detections predicted by the actor detector greedily when those have a high spatial affinity with boxes generated by the tracker. In practice, we use an overlap threshold of $0.7$. 
\vspace{-4mm}
\paragraph{Actor attention.}
Our video encoder corresponds to the convolutional stages of VGG16. We choose this base network for fair comparison with previous work based on deep learning models \cite{LiGJS16}.
Moreover, we only consider the RGB stream to encode the visual appereance of the videos in our experiments. The grid size for the bilinear interpolation of our Actor of Interest Pooling is $5\times5$. During training, our attention module focuses on the $k=12$ most relevant actors out of 20 actor tubes for classifying the video. Due to memory constraints, we employ segment partition introduced in \cite{TSN2016ECCV} to allocate more than one video per batch. For each video, we analyze 16 equally spaced frames each time.
We initialize the weights of our action classification model with Xavier technique, and our video encoder with the weights from a VGG16 model pre-trained on ILSVCR-12 \cite{RussakovskyDSKS15}.
We train our entire actor attention module end-to-end using SGD with momemtum in a single GPU with a batch size of four videos.

%------------------------------------------------------------------------
%------------------------------------------------------------------------
\subsection{Quality of Actor Proposals}
\vspace{-1mm}
We start by assessing the quality of our actor proposals. 
\vspace{-5mm}
\paragraph{Actor linking versus Viterbi linking.}
In our fist experiment we compare our actor linking with the more traditional Viterbi linking \cite{gkioxariCVPR15actionTubes} for the generation of actor tubes from the predictions of our actor detector. Our actor linking achieves an improvement in Recall of +19.9\% and +21.6\% for 0.2 and 0.5 IoU in THUMOS13. We attribute these good results to the capability of the similarity-based matching to accomodate for deformation of the actor, that the Viterbi linking approach is unable to fix. Previous approaches \cite{gkioxariCVPR15actionTubes,klaser2012human, Saha_bmvc16} employ supervision at the level of boxes and length of the tubes to overcome this issue. This clearly limits their application under the weakly-supervised setup evaluated in this work. We conclude that actor linking, aided by similarity-based matching, plays a crucial role towards spatiotemporal action localization with weak supervision.
\begin{table}[!ht]
\centering
\scalebox{0.90}{
\begin{tabular}{lccc}
\toprule
& \multicolumn{1}{c}{\textbf{UCF-Sports}} & \multicolumn{2}{c} {\textbf{THUMOS13}} \\
& IoU=0.5  & IoU=0.5 & IoU=0.1\\
\cmidrule(r){2-2} \cmidrule(rr){3-4}
Weinzaepfel \etal \cite{WeinzaepfelHS15} & 98.8  & - & - \\
Zhu \etal \cite{ZhuVL17} & 96.8  & 61.4 & -   \\
\midrule
Yu and Yuan \cite{yuCVPR15fasttubes} & - & - & 54.5 \\
van Gemert \etal \cite{Gemert_BMVC15} & 89.4 & 35.5 & - \\
Jain \etal \cite{Jain2017} & 91.9  & 32.8 &  -  \\
\textit{This paper}   & \textbf{93.6} & \textbf{43.9} & \textbf{88.7}\\
\bottomrule
\end{tabular}
}
\vspace{1mm}
\caption{Comparison of action proposals in terms of Recall. Weinzaepfel \etal \cite{WeinzaepfelHS15} and Zhu \etal \cite{ZhuVL17} use video supervision from action boxes and action labels, while the rest do not use any video supervision. Our actor proposals achieve better Recall compared to previous unsupervised and weakly-supervised methods.}
\label{proposal_recall_table}
\vspace{-6mm}
\end{table}
\begin{figure*}[!th]
\centering
    \includegraphics[width=0.9\textwidth]{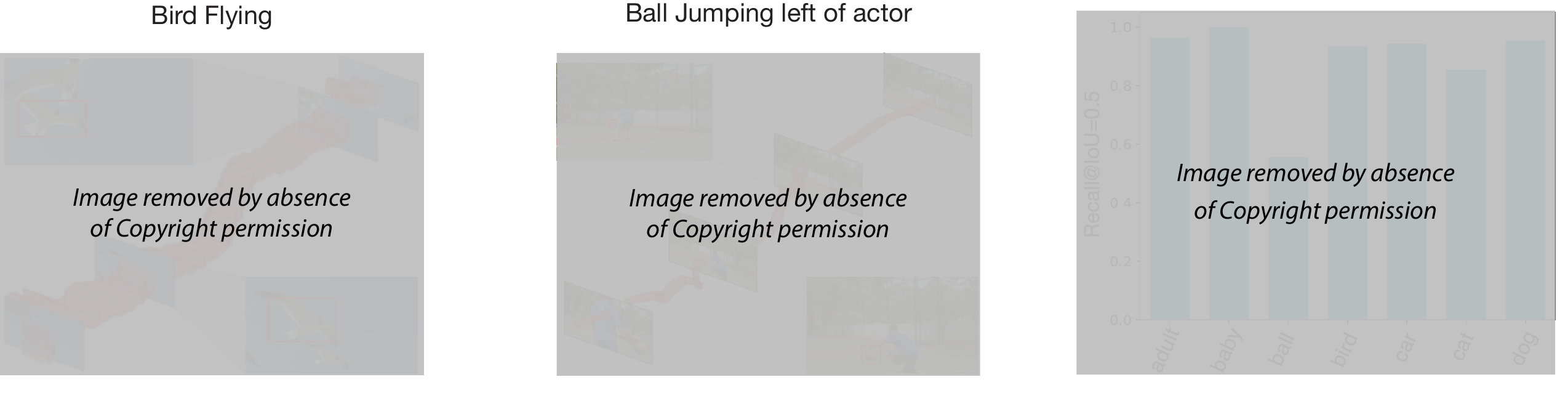}
%   \fbox{\rule{0pt}{1.5in} \rule{0.9\linewidth}{0pt}}
%\vspace{-10pt}
\caption{Our actor proposals also outline non-human actors.
We appreciate consistently high recall values in both articulated and rigid objects such bird, car, cat, and dog.
Interestingly, our approach also works, in less proportion thought,  for the \emph{Ball} categories.
This is understandable due to its common shape and small size, which invite many occlusions.
Please visit \url{https://goo.gl/hzywqm} for the actual qualitative and quantitative results.}
\label{fig:qualitative-a2d}
\vspace{-5pt}
\end{figure*}

\paragraph{Actor proposals versus others.}
Table \ref{proposal_recall_table} compares our actor proposals with previous supervised and unsupervised methods for action proposals. Compared to the action proposals by Yu and Yuan \cite{yuCVPR15fasttubes}, our approach achieves as substantial improvement of +34.2\% in terms of Recall on THUMOS13. This result evidences the benefit of our deep learning based actor detection and linking scheme, over traditional actor detection and linking. Interestingly, our approach improves upon previous unsupervised work by +1.7\% and +8.4\% in terms of Recall on UCF-Sports and THUMOS13, respectively. These methods \cite{Jain2017, Gemert_BMVC15} are based on grouping techniques over low-level primitives such as color and motion, which reaffirms our intuition about the relevance of actors as a strong cue for the localization of the actions.

The state of the art approaches for the generation of action proposals \cite{ZhuVL17, WeinzaepfelHS15} are fully supervised based on a mix of convolutional and recurrent stages and supervised instance level tracking, respectively. Although these works generate action proposals of better quality, they do so at the cost of a significant amount of additional supervision. This dependence on extra annotations limits their scalability potential.

Figure \ref{fig:proposal-comparison} illustrates the recall of our actors proposals for a varying number of proposal in comparison with previous unsupervised approaches. It provides further evidence on the quality of our proposals, especially when considering only a limited number of proposals.

\paragraph{Non-human actor proposals.} We also analyze the quality of our actor proposals for generating spatiotemporal tubes for non-human actors.
Figure \ref{fig:qualitative-a2d} summarizes our findings in this area. Please visit \url{https://goo.gl/hzywqm} for more details of the results of this experiment.
%The leftmost example corresponds to a successfull case of our approach.
Our method is able to generate proposals for highly articulated objects like \textit{birds} and \textit{cats}.
A common failure case of our approach in the dataset used is for \textit{ball} actor.
In most of the cases, the ball changes significantly in appeareance during the execution of the action with interleaving full occlussions by a human. From our quantitative analysis, we appreciate that the actor with the highest recall is \textit{baby}, which is not directly represented in the training set of our actor detector.
Similarly, the results reveals that the gap in recall at 0.5 IoU between adult actors and animal actors like \textit{bird} and \textit{dog} is at most +2\%. In general, for all the actors except for \textit{ball} the recall for all the actors using at most 50 actor proposals is greater than 85\%. Therefore, we conclude that our method is general and applicable to human and non-human actors.
%------------------------------------------------------------------------
%------------------------------------------------------------------------
\subsection{Spatiotemporal Action Localization}
Our approach not only generates candidate actor proposals, but it also predicts the associated action class. To evaluate its performance, we employ the evaluation setup typical for action localization using mean average precision (mAP) over all the classes in the dataset, given an overlap threshold of 0.2 (THUMOS13) and 0.5 (UCF-Sports).

Table \ref{table:action-localization} compares multiple approaches with a varying degree of supervision. Based on these results, we note that our actor supervision approach achieves the state-of-the-art among all weakly supervised approaches. It improves upon them by +7.8\% and +10.4\% on the THUMOS13 and UCF-Sports benchmarks, respectively.
We attribute that to an appropriate architectural design that eases the localization of the actions by the use of actor supervision. Compared to \cite{MettesSC17}, our approach gives more relevance to the actors during the localization stage instead of using them as cues to improve the ranking of existing action proposals. We hypothesize that our attention mechanism is more effective than the one in \cite{LiGJS16}, because actors are a more powerful cue for guiding the localization of actions than individual pixels.

Actor-supervision also outperforms several approaches with varying levels of supervision on the challenging THUMOS13 benchmark \cite{yuCVPR15fasttubes, Gemert_BMVC15, MettesECCV16}. Compared to \cite{WeinzaepfelHS15}, our visual representation is limited to the RGB video stream, which clearly leaves room for improvement by richer features with others including optical flow.

\begin{table*}[!t]
\centering
\begin{tabular}{lccccrr}
\toprule
& \multicolumn{4}{c}{\textbf{Action Supervision}} & \multicolumn{1}{c} {\textbf{THUMOS13}} & \multicolumn{1}{c} {\textbf{UCF-Sports}}\\
& Boxes & Segments & Points & Labels & mAP@0.2 & mAP@0.5 \\  
\cmidrule(lr){2-5} \cmidrule(lr){6-6} \cmidrule(lr){7-7}
Kalogeiton \etal \cite{kalogeiton17iccv} & \checkmark & \checkmark & & \checkmark & 77.2 & 92.7 \\
Saha \etal \cite{GSingh_iccv17_online}   & \checkmark & \checkmark & & \checkmark & 73.5 & - \\
Hou \etal \cite{HouCS_iccv17} & \checkmark & & & \checkmark & 47.1 & 86.7 \\
Jain \etal \cite{Jain2017}               & \checkmark & \checkmark & & \checkmark & 48.1 & - \\
Weinzaepfel \etal \cite{WeinzaepfelHS15} & \checkmark & \checkmark & & \checkmark & 46.8 & 90.5 \\
van Gemert \etal \cite{Gemert_BMVC15}    & \checkmark & \checkmark & & \checkmark & 37.8 & - \\
Yu \etal \cite{yuCVPR15fasttubes}        & \checkmark & \checkmark & & \checkmark & 26.5 & - \\
Mettes \etal \cite{MettesECCV16}         & & & \checkmark & \checkmark & 34.8 & - \\
\midrule
Mettes \etal \cite{MettesSC17}           & & & & \checkmark & 37.4 & 37.8 \\
Li \etal \cite{LiGJS16}                  & & & & \checkmark & 37.7 & -    \\
Cinbis \etal \cite{cinbis2014multi} (from \cite{MettesECCV16}) & & & & \checkmark & 13.6 & -    \\
Sharma \etal \cite{sharma2015action} (from \cite{LiGJS16}) & & & & \checkmark & 5.5 & -    \\
\textit{This paper}                      & & & & \checkmark & \textbf{45.5} & \textbf{48.2} \\
\bottomrule
\end{tabular}
\vspace{10pt}
\caption{Comparison of spatiotemporal action localization approaches with decreasing amount of supervision. The top half shows supervised approaches, whereas the bottom half shows weakly-supervised approaches relying on action class labels only. Our architecture guided by actor-supervision achieves state-of-the-art performance among the weakly-supervised approaches.}
\label{table:action-localization}
\end{table*}

\begin{figure*}[!h]
\centering
  \includegraphics[width=0.9\textwidth]{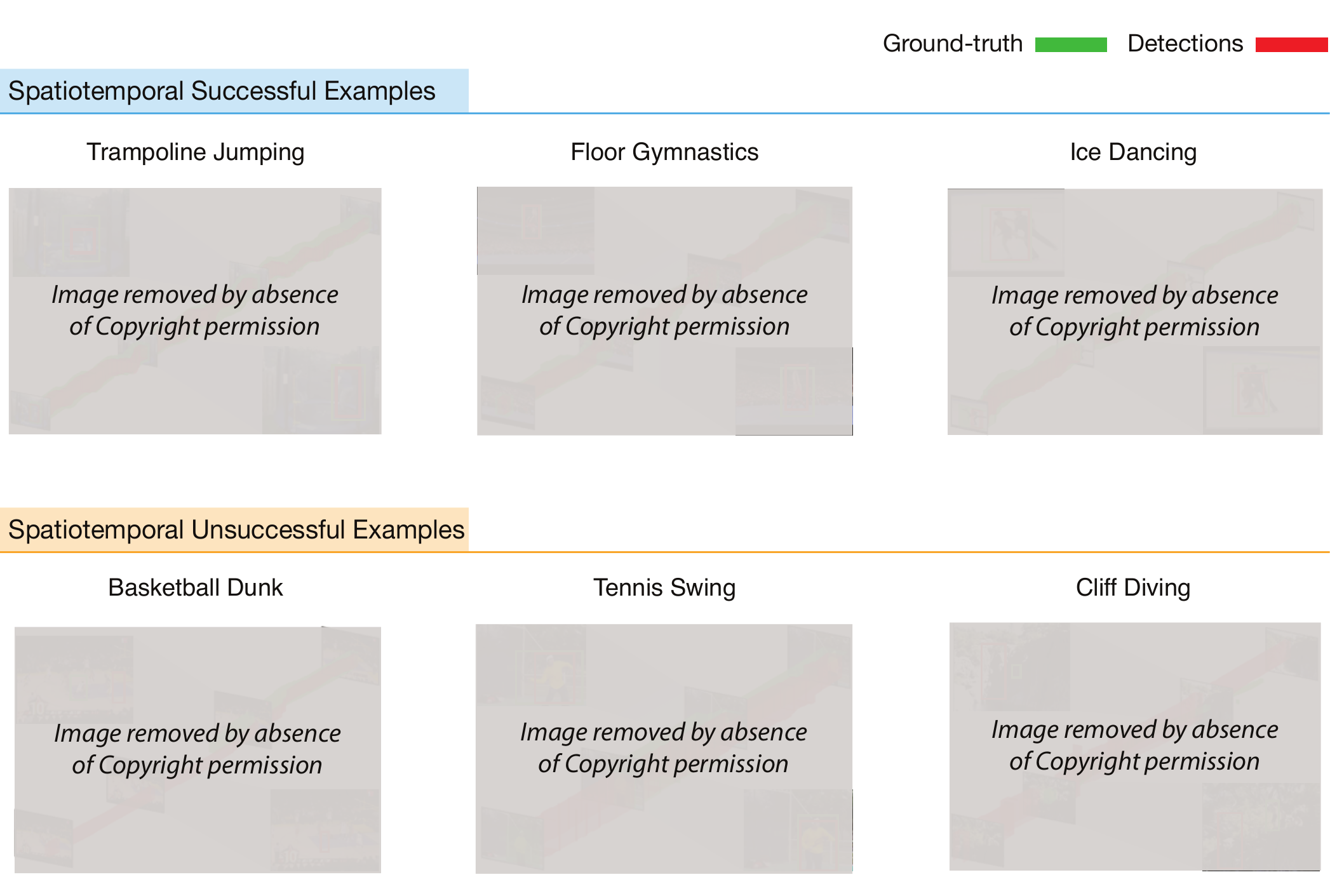}
\caption{Qualitative results on the THUMOS13 dataset: Top row shows three  successful cases by visualizing the ground-truth and action tubes as well as two highlighted frames. These include action sequences that have deformations of actor as well as multiple actors with complex background. Bottom row visualizes three failed cases which show that crowded background, occlusions and temporally untrimmed action sequences are the most challenging scenarios.
Please visit \url{https://goo.gl/hzywqm} for the actual qualitative results.}
\label{fig:qualitative-results}
\end{figure*}

\paragraph{Comparison with the state-of-the-art.}
The state-of-the-art in action localization is dominated by fully-supervised approaches resembling conv-net architectures well established for generic object detection \cite{kalogeiton17iccv, GSingh_iccv17_online}, which require strong levels of supervision. These approaches are unable to be trained in the weakly supervised regime presented in this paper. 
Interestingly, our approach not only considerably outperforms other weakly supervised methods but it also has an edge over some of the supervised approaches. 
To the best of our knowledge, our architecture is the first fully end-to-end approach tackling the problem of spatiotemporal localization of actions without resorting to action box supervision. Considering the poor scalability of fully-supervised approaches and the tremendous amount of progress in object detection, we envision that our work can inspire the community to seek other forms of supervision during the design and adaption of deep representations for localizing actions in videos.
Figure \ref{fig:qualitative-results} illustrates action localization results of our approach on the THUMOS13 benchmark.

%------------------------------------------------------------------------
%%%%%%%%%%%%%%%%%%%%%%%%%%%%%%%%%%%%%%%%%%%%%%%%%%%%%%%%%%%%%%%%%%%%%%%%%
%%%%%%%%%%%%%%%%%%%%%%%%%%%%%%%%%%%%%%%%%%%%%%%%%%%%%%%%%%%%%%%%%%%%%%%%%
%------------------------------------------------------------------------

\section{Conclusion}
This paper introduces a weakly supervised approach for the spatiotemporal localization of actions in video, driven by actor supervision. We show that exploiting the inherent compositionality of actions, in terms of transformations of \textit{actors}, disregards the dependence on spatiotemporal annotations of the training videos. In the proposal generation step, we introduce actor supervision in the form of an actor detector and similarity-based matching to locate the action in the video as a set of actor proposals. Then, our proposed actor attention learns to classify and rank these actor proposals for a given action class. This step also does not require any per frame box-level annotations. Our approach outperforms the state-of-the-art among weakly supervised works and even achieves results that are better or competitive to some of the fully-supervised methods.

\paragraph{Acknowledgment}
This work was supported by the King Abdullah University of Science and Technology
(KAUST) Office of Sponsored Research. We thank the team members of the IVUL and Morpheus from KAUST and Qualcomm Technologies, Inc. for helpful comments and discussion. In particular, we appreciate the support of Amirhossein Habibian during the implemention of our Actor Linking.

\begin{figure*}[!ht]
\begin{center}
  \centering
  \includegraphics[width=0.98\textwidth]{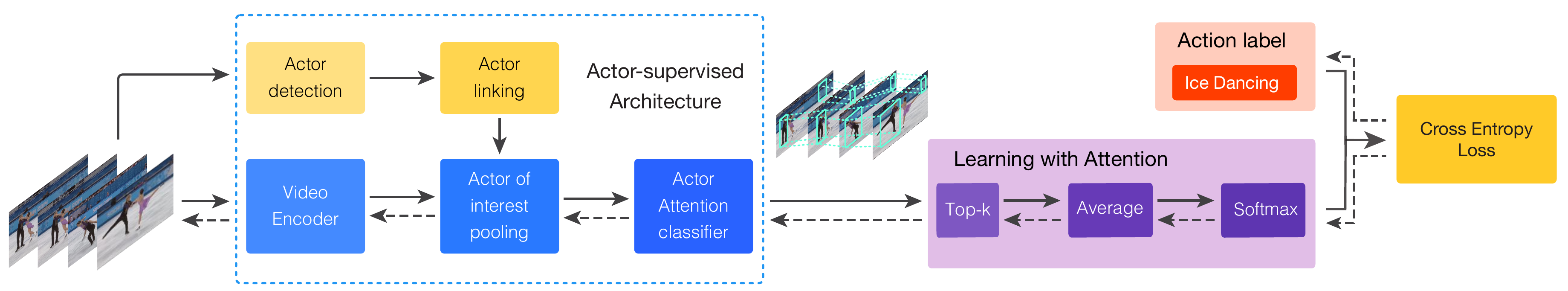}
\end{center}%
\caption{Our actor-supervised architecture is the first fully end-to-end approach for weakly-supervised spatiotemporal action localization.
The block diagram illustrates all the operations of our approach during training.
The dashed lines represent the flow of data during backpropagation}
\label{fig:pipeline_learning}
\end{figure*}

{\small
\bibliographystyle{ieee}
\bibliography{egbib}

\begin{thebibliography}{10}\itemsep=-1pt

\bibitem{bertinetto2016fully}
L.~Bertinetto, J.~Valmadre, J.~F. Henriques, A.~Vedaldi, and P.~H.~S. Torr.
\newblock Fully-convolutional siamese networks for object tracking.
\newblock In {\em ECCV 2016 Workshops}, 2016.

\bibitem{chen2015action}
W.~Chen and J.~Corso.
\newblock Action detection by implicit intentional motion clustering.
\newblock In {\em ICCV}, 2015.

\bibitem{cinbis2014multi}
R.~G. Cinbis, J.~Verbeek, and C.~Schmid.
\newblock Multi-fold mil training for weakly supervised object localization.
\newblock In {\em CVPR}, 2014.

\bibitem{girshickICCV15fastrcnn}
R.~Girshick.
\newblock Fast r-cnn.
\newblock In {\em ICCV}, 2015.

\bibitem{gkioxariCVPR15actionTubes}
G.~Gkioxari and J.~Malik.
\newblock Finding action tubes.
\newblock In {\em CVPR}, 2015.

\bibitem{HouCS_iccv17}
R.~Hou, C.~Chen, and M.~Shah.
\newblock Tube convolutional neural network (t-cnn) for action detection in
  videos.
\newblock In {\em ICCV}, 2017.

\bibitem{JaderbergSZK15}
M.~Jaderberg, K.~Simonyan, A.~Zisserman, and K.~Kavukcuoglu.
\newblock Spatial transformer networks.
\newblock In {\em NIPS}, 2015.

\bibitem{Jain_cvpr14}
M.~Jain, J.~van Gemert, H.~J{\'e}gou, P.~Bouthemy, and C.~Snoek.
\newblock Action localization with tubelets from motion.
\newblock In {\em CVPR}, 2014.

\bibitem{Jain2017}
M.~Jain, J.~van Gemert, H.~J{\'e}gou, P.~Bouthemy, and C.~Snoek.
\newblock Tubelets: Unsupervised action proposals from spatiotemporal
  super-voxels.
\newblock {\em IJCV}, 124(3):287--311, 2017.

\bibitem{Jain15iccv}
M.~Jain, J.~van Gemert, T.~Mensink, and C.~Snoek.
\newblock Objects2action: Classifying and localizing actions without any video
  example.
\newblock In {\em ICCV}, 2015.

\bibitem{densecap}
J.~Johnson, A.~Karpathy, and L.~Fei-Fei.
\newblock Densecap: Fully convolutional localization networks for dense
  captioning.
\newblock In {\em CVPR}, 2016.

\bibitem{kalogeiton17iccv}
V.~Kalogeiton, P.~Weinzaepfel, V.~Ferrari, and C.~Schmid.
\newblock {Action Tubelet Detector for Spatio-Temporal Action Localization}.
\newblock In {\em {ICCV}}, 2017.

\bibitem{klaser2012human}
A.~Kl{\"a}ser, M.~Marsza{\l}ek, C.~Schmid, and A.~Zisserman.
\newblock Human focused action localization in video.
\newblock In {\em Trends and Topics in Computer Vision}, pages 219--233, 2012.

\bibitem{tian_iccv11}
T.~Lan, Y.~Wang, and G.~Mori.
\newblock Discriminative figure-centric models for joint action localization
  and recognition.
\newblock In {\em ICCV}, 2011.

\bibitem{LiGJS16}
Z.~Li, K.~Gavrilyuk, E.~Gavves, M.~Jain, and C.~Snoek.
\newblock Videolstm convolves, attends and flows for action recognition.
\newblock {\em CVIU}, 2017.

\bibitem{LinMBHPRDZ14}
T.~Lin, M.~Maire, S.~J. Belongie, J.~Hays, P.~Perona, D.~Ramanan,
  P.~Doll{\'{a}}r, and C.~L. Zitnick.
\newblock Microsoft {COCO:} common objects in context.
\newblock In {\em ECCV}, 2014.

\bibitem{SSD_eccv16}
W.~Liu, D.~Anguelov, D.~Erhan, C.~Szegedy, S.~Reed, C.-Y. Fu, and A.~Berg.
\newblock Ssd: Single shot multibox detector.
\newblock In {\em ECCV}, 2016.

\bibitem{MaZIS13}
S.~Ma, J.~Zhang, N.~Ikizler-Cinbis, and S.~Sclaroff.
\newblock Action recognition and localization by hierarchical space-time
  segments.
\newblock In {\em ICCV}, 2013.

\bibitem{MettesICCV17}
P.~Mettes and C.~Snoek.
\newblock Spatial-aware object embeddings for zero-shot localization and
  classification of actions.
\newblock In {\em ICCV}, 2017.

\bibitem{MettesSC17}
P.~Mettes, C.~Snoek, and S.~Chang.
\newblock Localizing actions from video labels and pseudo-annotations.
\newblock In {\em BMVC}, 2017.

\bibitem{MettesECCV16}
P.~Mettes, J.~van Gemert, and C.~Snoek.
\newblock Spot on: Action localization from pointly-supervised proposals.
\newblock In {\em ECCV}, 2016.

\bibitem{oneataECCV14spatemprops}
D.~Oneata, J.~Revaud, J.~Verbeek, and C.~Schmid.
\newblock Spatio-temporal object detection proposals.
\newblock In {\em ECCV}, 2014.

\bibitem{marianICCV2015unsupervisedTube}
M.~Puscas, E.~Sangineto, D.~Culibrk, and N.~Sebe.
\newblock Unsupervised tube extraction using transductive learning and dense
  trajectories.
\newblock In {\em ICCV}, 2015.

\bibitem{faster_rcnn_nips15}
S.~Ren, K.~He, R.~Girshick, and J.~Sun.
\newblock Faster r-cnn: Towards real-time object detection with region proposal
  networks.
\newblock In {\em NIPS}, 2015.

\bibitem{RodriguezCVPR2008}
M.~D. Rodriguez, J.~Ahmed, and M.~Shah.
\newblock Action {MACH}: a spatio-temporal maximum average correlation height
  filter for action recognition.
\newblock In {\em CVPR}, 2008.

\bibitem{RussakovskyDSKS15}
O.~Russakovsky, J.~Deng, H.~Su, J.~Krause, S.~Satheesh, S.~Ma, Z.~Huang,
  A.~Karpathy, A.~Khosla, M.~S. Bernstein, A.~C. Berg, and F.~Li.
\newblock Imagenet large scale visual recognition challenge.
\newblock {\em IJCV}, 115(3):211--252, 2015.

\bibitem{Saha_iccv17_AMTnet}
S.~Saha, G.~Singh, and F.~Cuzzolin.
\newblock Amtnet: Action-micro-tube regression by end-to-end trainable deep
  architecture.
\newblock In {\em ICCV}, 2017.

\bibitem{Saha_bmvc16}
S.~Saha, G.~Singh, M.~Sapienza, P.~Torr, and F.~Cuzzolin.
\newblock Deep learning for detecting multiple space-time action tubes in
  videos.
\newblock In {\em BMVC}, 2016.

\bibitem{GSingh_iccv17_online}
S.~Saha, G.~Singh, M.~Sapienza, P.~Torr, and F.~Cuzzolin.
\newblock Online real-time multiple spatiotemporal action localisation and
  prediction.
\newblock In {\em ICCV}, 2017.

\bibitem{sharma2015action}
S.~Sharma, R.~Kiros, and R.~Salakhutdinov.
\newblock Action recognition using visual attention.
\newblock In {\em NIPS workshop}, 2015.

\bibitem{SivaBMVC11}
P.~Siva and T.~Xiang.
\newblock Weakly supervised action detection.
\newblock In {\em BMVC}, 2011.

\bibitem{SmeuldersCCCDS14}
A.~W.~M. Smeulders, D.~M. Chu, R.~Cucchiara, S.~Calderara, A.~Dehghan, and
  M.~Shah.
\newblock Visual tracking: An experimental survey.
\newblock {\em TPAMI}, 36(7):1442--1468, 2014.

\bibitem{SoomroS17}
K.~Soomro and M.~Shah.
\newblock Unsupervised action discovery and localization in videos.
\newblock In {\em ICCV}, 2017.

\bibitem{SoomroZ2014}
K.~Soomro and A.~R. Zamir.
\newblock {\em Action Recognition in Realistic Sports Videos}, pages 181--208.
\newblock Springer International Publishing, 2014.

\bibitem{ucf101}
K.~Soomro, A.~R. Zamir, and M.~Shah.
\newblock {UCF101:} {A} dataset of 101 human actions classes from videos in the
  wild.
\newblock {\em CoRR}, 2012.

\bibitem{tao2016sint}
R.~Tao, E.~Gavves, and A.~W.~M. Smeulders.
\newblock Siamese instance search for tracking.
\newblock In {\em CVPR}, 2016.

\bibitem{Tran:nips12}
D.~Tran and J.~Yuan.
\newblock Max-margin structured output regression for spatio-temporal action
  localization.
\newblock In {\em NIPS}, 2012.

\bibitem{Jasper:selective}
J.~Uijlings, K.~van~de Sande, T.~Gevers, and A.~Smeulders.
\newblock Selective search for object recognition.
\newblock {\em IJCV}, 104(2):154--171, 2013.

\bibitem{Gemert_BMVC15}
J.~van Gemert, M.~Jain, E.~Gati, and C.~Snoek.
\newblock {APT}: Action localization proposals from dense trajectories.
\newblock In {\em BMVC}, 2015.

\bibitem{TSN2016ECCV}
L.~Wang, Y.~Xiong, Z.~Wang, Y.~Qiao, D.~Lin, X.~Tang, and L.~{Van Gool}.
\newblock Temporal segment networks: Towards good practices for deep action
  recognition.
\newblock In {\em ECCV}, 2016.

\bibitem{WeinzaepfelHS15}
P.~Weinzaepfel, Z.~Harchaoui, and C.~Schmid.
\newblock {Learning to track for spatio-temporal action localization}.
\newblock In {\em ICCV}, 2015.

\bibitem{XuHsXiCVPR2015}
C.~Xu, S.-H. Hsieh, C.~Xiong, and J.~{Corso}.
\newblock Can humans fly? {Action} understanding with multiple classes of
  actors.
\newblock In {\em {CVPR}}, 2015.

\bibitem{yuCVPR15fasttubes}
G.~Yu and J.~Yuan.
\newblock Fast action proposals for human action detection and search.
\newblock In {\em CVPR}, 2015.

\bibitem{ZhuVL17}
H.~Zhu, R.~Vial, and S.~Lu.
\newblock {TORNADO}: A spatio-temporal convolutional regression network for
  video action proposal.
\newblock In {\em ICCV}, 2017.

\bibitem{Zitnick2014}
L.~Zitnick and P.~Doll{\'a}r.
\newblock Edge boxes: Locating object proposals from edges"s.
\newblock In {\em ECCV}, 2014.

\end{thebibliography}
}
%\cleardoublepage

%\twocolumn[{%
%\renewcommand\twocolumn[1][]{#1}%
%\begin{center}
%  \centering
%  \includegraphics[width=0.98\textwidth]{figure2_backprop.pdf}
%\end{center}%
%\captionof{Figure 7. }{Our actor-supervised architecture is the first fully end-to-end approach for weakly-supervised spatiotemporal action localization.
%The block diagram illustrates all the operations of our approach during training.
%The dashed lines represent the flow of data during backpropagation}
%\label{fig:pipeline_learning}
%\vspace{5mm}
%}]

\section{Appendix}

We complement our work with the following items:

\begin{itemize}
    \item the pseudocode of our approach for generating actor proposals (Section \ref{algorithm-section}).
    \item Details about the training of our actor-supervised architecture (Section \ref{details-section}).
\end{itemize}

\subsection{Actor Proposals} \label{algorithm-section}

The algorithm \ref{aa-algorithm} describes all the interactions between the inner blocks involved for the generation of our actor proposals, described in the paper.

\begin{algorithm}[!h]
 \begin{algorithmic}[1]
 \State \textbf{Input:} maximum number of proposals $N$ % time limit $\tau_{max}$, 
 \State \textbf{Output:} $\mathcal{T}$
 \State $\mathcal{D} \leftarrow$ run \texttt{actor detector} over all frames
 \State $\mathcal{T} \leftarrow \emptyset$
 \State $i \leftarrow 0$
 \While{$\mathcal{D} \neq \emptyset \land i < N$} % \land time $\leq \tau_{max}$
    \State $b_i \leftarrow $ \texttt{select} actor with highest score from $\mathcal{D}$
    \State $\mathcal{B}_i \leftarrow $ \texttt{actor tracker} tracks $b_i$ forward and backward throughout the video 
    \State \texttt{Push} $\mathcal{B}_i$ onto $\mathcal{T}$
    \State $\mathcal{D} \leftarrow$ \texttt{filter} actors in $\mathcal{D}$ with high similarity with boxes in $\mathcal{B}_i$
    \State $i \leftarrow i + 1$
 \EndWhile
 \end{algorithmic}
 \caption{Actor proposals generation}
 \label{aa-algorithm}
\end{algorithm}

\subsection{Actor-supervised architecture} \label{details-section}

\paragraph{Training details of our actor-attention stream}
We train our actor attention stream for 30 epochs annealing the learning rate by a factor of $0.25$ after five epochs.
We use a momentum factor of $0.95$ and an initial learning rate of $0.001$.
We only apply weight decay to the weights of our actor classification module with a factor of $1\times10^{-4}$.
As input pre-processing, we employ the segment based strategy suggested by \cite{TSN2016ECCV}. In our case, we randomly sample 16 frames uniformly spaced per video. Additionally, we apply a random horizontal flipping of all the sampled frames.
Finally, we normalize the input frames such that the intensity values lie on the range between $[-0.5, 0.5]$ using standard scaling with mean $[0.485, 0.456, 0.406]$ and standard deviation $[0.229, 0.224, 0.225]$ for the RGB channel.

\paragraph{Backpropagation details}
% Figure \ref{fig:pipeline_learning}
Figure 7 illustrates the modules involved during the training of our actor attention. These are (i) our video encoder; (ii) our actor of interect pooling; (iii) our actor attention classifier; (iv) top-k selection; (v) average among top-k scores per-class; and (vi) softmax activation.
Based on this diagram, we claim that our architecture is the first fully end-to-end work for weakly-supervised spatiotemporal action localization.
During training, we tune all the parameters of our video encoder. Our approach differs from the training procedure in \cite{LiGJS16} which trains a recurrent module on top of pre-computed features from the last convolutional block of VGG16.
%Finally, it is worth to mention that it is possible to fine tune our actor proposals stream with our architecture, but we leave this extension for future work.

\end{document}